%% file: worldtok_iclr2027.tex
\documentclass{article}

\usepackage{iclr2027_conference}
\usepackage{times}
\input{math_commands.tex}

\usepackage[utf8]{inputenc}
\usepackage[T1]{fontenc}
\usepackage{hyperref}
\usepackage{url}
\usepackage{booktabs}
\usepackage{amsfonts}
\usepackage{amssymb}
\usepackage{nicefrac}
\usepackage{microtype}
\usepackage{graphicx}
\usepackage{subcaption}
\usepackage{amsmath}
\usepackage{multirow}
\usepackage{makecell}
\usepackage[table]{xcolor}
\usepackage{afterpage}
\usepackage{placeins}

\iclrfinalcopy %

\title{DriveUniTok: A Unified Multimodal Tokenizer for Driving World-Action Models}

\author{%
\bfseries Jingbo Xu\textsuperscript{1,2}, Hanchao Leng\textsuperscript{2,$*$}, JiaHui Huang\textsuperscript{2}, Hanbing Li\textsuperscript{2}, Jingwei Zhao\textsuperscript{2},\\
\bfseries Kuiyuan Yang\textsuperscript{2}, Chen Gao\textsuperscript{1}, Guang Chen\textsuperscript{2}, Hangjun Ye\textsuperscript{2}, Yong Li\textsuperscript{1,$\dagger$}\\[0.5em]
\normalfont\textsuperscript{1}Tsinghua University \quad \textsuperscript{2}Xiaomi EV\\[0.4em]
\textsuperscript{$*$}Project leader. \quad \textsuperscript{$\dagger$}Corresponding author.
}

\begin{document}

\maketitle
\fancyhead{}
\renewcommand{\headrulewidth}{0pt}
\begin{abstract}
World models are emerging as a fundamental component of policy learning for real-world agents, enabling agents to predict future world states and plan actions within a learned representation of the dynamic environment. Existing approaches, however, often model appearance, metric geometry represented by projected LiDAR depth, and ego motion either partially or independently, overlooking the fact that these modalities are different projections of a shared underlying world state. This separation can lead to information loss and cross-modal inconsistency, ultimately limiting downstream performance. Motivated by this unified view of the world, we propose \textbf{DriveUniTok}, a unified multimodal tokenizer for driving world-action models (WAMs) that encodes these modalities in a single decodable continuous latent. We freeze DriveUniTok and adapt a pretrained diffusion model, \textbf{DriveUniWAM}, to predict future unified latents; the tokenizer then decodes them into future appearance, projected LiDAR depth, and ego trajectories. Experiments on NAVSIM demonstrate that our unified design achieves better future prediction and planning performance at lower downstream computational cost than separate-stream alternatives. With reinforcement learning post-training, DriveUniWAM achieves a PDMS of 91.54, reaching state-of-the-art performance among the evaluated WAM-based planners.

\end{abstract}

\input{sections/introduction}
\input{sections/related_work}

\input{sections/methodology}
\input{sections/experiments}
\input{sections/conclusion}
\input{sections/ai_use_statement}
\input{sections/reproducibility_statement}

\input{sections/references}
\input{sections/supplementary}

\end{document}

%% file: math_commands.tex
\usepackage{amsmath,amsfonts,bm}

\def\eqref#1{equation~\ref{#1}}

\def\1{\bm{1}}

\DeclareMathAlphabet{\mathsfit}{\encodingdefault}{\sfdefault}{m}{sl}
\SetMathAlphabet{\mathsfit}{bold}{\encodingdefault}{\sfdefault}{bx}{n}

%% file: sections/introduction.tex
\section{Introduction}

Driving world models initially served as scene simulators and data engines, forecasting future observations for data augmentation and policy training. More recently, generative models of future observations have been integrated into policy backbones that jointly model scene evolution and ego trajectories \citep{zhang2025epona,zhao2025pwm,xia2026drivelaw,shi2026drivewam}. In this setting, the latent representation is no longer merely an interface for visual reconstruction; it is the space in which the model captures the information required for both future prediction and planning.

\begin{figure}[t]
\centering
\includegraphics[width=0.9\textwidth]{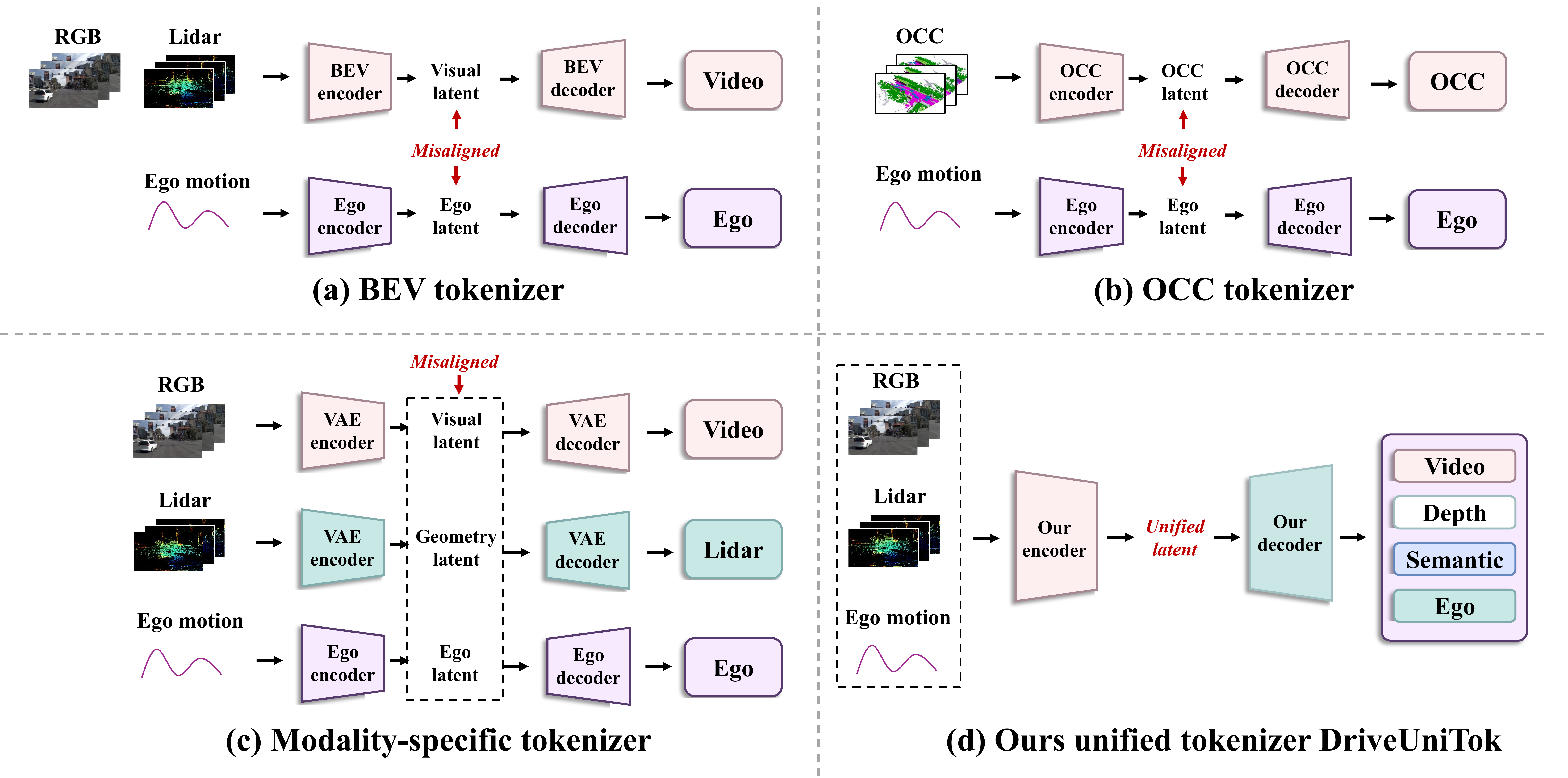}
\caption{Comparison of latent representations for world-action modeling. (a) BEV tokenizers fuse RGB and LiDAR into a visual latent, while (b) Occupancy tokenizers represent the scene with an occupancy latent; both keep ego motion separate. (c) Modality-specific tokenizers encode RGB, LiDAR, and ego motion into separate streams. (d) DriveUniTok jointly encodes RGB, LiDAR-derived geometry, and ego motion into one unified latent.}
\label{fig:fusion_comparison}
\end{figure}

Existing world-action models, however, expose asymmetric representation interfaces. Most inherit a pretrained RGB video autoencoder whose latent primarily captures appearance, while geometry is introduced through auxiliary supervision and ego pose remains outside the latent, where it is handled by a separate conditioning stream or action head \citep{wan2025wan,kong2024hunyuanvideo,agarwal2025cosmos}. Driving tokenizers fuse camera and LiDAR observations into scene-centric bird's-eye-view latents, but still leave ego pose outside the scene representation \citep{zhang2024bevworld,chen2025bevvae,zhuo2026drivetok,liao2025i2world}. Cross-modal and latent-action tokenizers incorporate actions alongside observations, yet commonly retain modality-specific streams or align them through a designated anchor or intermediary representation \citep{lu2026atoken,bachmann2024fourm21,chen2025villax,fan2025xr1,chen2026unit}. Figure~\ref{fig:fusion_comparison} compares these representations according to whether appearance, metric geometry, and ego pose reside in the same decodable latent.

This asymmetry forces the downstream predictor to learn cross-modal correspondence together with temporal evolution, which can make future-scene and ego-trajectory predictions inconsistent. Therefore, a central question arises: Should visual appearance, metric geometry, and ego-motion be represented within a unified tokenization framework? DriveUniTok addresses this question by encoding RGB appearance, projected LiDAR geometry, and clip-relative ego pose in one jointly decodable latent.

\raggedbottom
\afterpage{\flushbottom}

To construct such a representation, we propose \textbf{DriveUniTok}, a query-based multimodal tokenizer. Shared learnable queries aggregate appearance features from RGB, metric geometry features from projected LiDAR depth, and clip-relative ego-pose features into a common sequence of per-frame multimodal tokens. A causal temporal compressor then maps this sequence into a compact continuous unified latent. All three modalities are reconstructed from the shared bottleneck through modality-specific readouts, without using a privileged anchor modality. In addition, a training-only semantic decoder is supervised by representations from a vision foundation model, encouraging the unified latent to preserve high-level scene semantics alongside metric geometry and pose.

To evaluate whether this representation supports joint forecasting and planning, we freeze DriveUniTok and adapt a pretrained diffusion model into \textbf{DriveUniWAM}, using the same downstream backbone and training protocol for unified and separate representations. The Wan-initialized flow-matching transformer predicts future unified latents under block-causal attention, conditioned on DriveUniTok-encoded historical observations and clip-relative ego poses together with a navigation signal. Future RGB, depth, and ego trajectories are decoded from the same predicted latent evolution rather than generated through independent observation and action pathways. Because supervised flow matching fits recorded futures but does not directly optimize the NAVSIM planning metric, we additionally post-train DriveUniWAM with GRPO against the PDM reward while regularizing it toward the supervised reference model.

Our key contributions are as follows:
\begin{itemize}
    \item We propose DriveUniTok, a query-based tokenizer that encodes RGB appearance, projected LiDAR geometry, and clip-relative ego pose in a unified continuous latent.
    \item We develop DriveUniWAM, a diffusion-based WAM that forecasts this latent for joint future-scene prediction and ego-trajectory planning.
    \item We show on NAVSIM that the unified design achieves stronger prediction and planning at lower downstream computational cost than separate-stream controls, with DriveUniWAM reaching state-of-the-art performance at 91.54 PDMS after reinforcement learning post-training.
\end{itemize}

%% file: sections/related_work.tex
\section{Related Work}

\subsection{General and Driving Tokenizers}

For visual tokenizers, discrete codebooks are commonly paired with autoregressive or masked token prediction \citep{van2017neural,yu2024magvitv2}, whereas continuous latents are often used with diffusion and flow-matching models. Large video VAEs provide high-fidelity latent spaces for RGB generation \citep{wan2025wan,kong2024hunyuanvideo,yang2025cogvideox,agarwal2025cosmos}, and quantization and model design materially affect representation quality \citep{luo2024openmagvit2,yu2024magvitv2,wang2024omnitokenizer,tang2024vidtok,wang2025larp,guo2026vrae}. General-purpose multimodal tokenizers place vision, language, audio, and other signals in a shared token space through semantic or contrastive objectives \citep{lu2026atoken,bachmann2024fourm21,lu2023unifiedio,lu2024unifiedio2}. These tokenizers, however, are optimized for general visual generation or cross-modal alignment, and do not explicitly capture the metric geometry or ego motion required for driving scenes. Driving-specific tokenizers instead focus on scene representations. BEV-VAE and DriveTok construct tokens from multiview observations \citep{chen2025bevvae,zhuo2026drivetok}, while OccWorld and $I^{2}$-World tokenize structured 3D occupancy for world modeling and 4D forecasting \citep{zheng2024occworld,liao2025i2world}. Moving beyond camera- or occupancy-centric scene tokens, BEVWorld fuses surround-view images and LiDAR into a BEV latent that can be rendered back to both modalities, but the latent still describes the scene without encoding ego pose \citep{zhang2024bevworld}. These limitations motivate a unified tokenizer that keeps appearance, metric geometry, and ego pose jointly decodable.

\subsection{Driving Planners and World-Action Models}

End-to-end driving systems jointly optimize perception and planning, from feature fusion and waypoint regression \citep{chitta2022transfuser} and planning-centric perception and prediction \citep{hu2023uniad} to anchored trajectory sets refined by truncated diffusion \citep{liao2025diffusiondrive} or autoregressive mixture-of-experts decoding \citep{feng2025artemis}. World-model-assisted planners expose future-scene information through future rollouts, predicted BEV features, or world-objective constraints on the planning latent \citep{li2025wote,zhang2025seerdrive,yang2026worldrft}. Vision-language-action planners add semantic priors and explicit reasoning, using either chain-of-thought reasoning with discrete action tokens or language-derived representations with diffusion planning and reward refinement \citep{zhou2025autovla,li2026recogdrive,li2026sgdrive}, and recent variants further incorporate generated futures through interleaved generation, latent conditioning, iterative replanning, or auxiliary supervision \citep{liu2026uniworldvla,li2026imagidrive,zuo2026vega,sheng2026explorevla,li2026drivevlaw0,zhou2026drivedreamerpolicy}. These planners use future information in different ways, but generally treat future-scene prediction and trajectory generation as separate modeling components.

World-action models instead couple the two within a generative model. Existing world-action models differ primarily in how predicted future observations interact with trajectory generation. Some use parallel diffusion heads over a shared spatiotemporal latent \citep{zhang2025epona} or attention over discrete future-frame tokens in an autoregressive sequence \citep{zhao2025pwm}, while others condition action transformers on cached video-denoising features \citep{xia2026drivelaw}, decode action chunks from a sampled future-video latent \citep{shi2026drivewam}, or autoregressively interleave scene and action tokens \citep{ye2025dap,chen2026drivinggpt}. While these designs couple future prediction and action generation, they do not generally provide a single decodable latent that jointly represents appearance, metric geometry, and ego motion.

In contrast, DriveUniTok unifies RGB appearance, projected LiDAR geometry, and clip-relative ego pose at the tokenization stage in one jointly decodable continuous latent. DriveUniWAM forecasts this latent for joint future-scene prediction and planning. Our contribution centers on unifying appearance, metric geometry, and ego motion at the representation level for world-action modeling.

%% file: sections/methodology.tex
\section{Methodology}

\subsection{Problem Setup}

A driving observation sequence $\{o_1,\dots,o_T\}$ contains $T=5$ consecutive frames sampled at 2\,Hz. Each observation $o_i=(I_i,D_i,p_i)$ consists of a front-camera image $I_i\in\mathbb{R}^{3\times H\times W}$, a sparse depth map $D_i\in\mathbb{R}^{1\times H\times W}$ obtained by projecting the LiDAR sweep onto the image plane of $I_i$, and an ego pose $p_i\in\mathbb{R}^{3}$ expressed relative to the first frame of the sequence, with $H\times W=288\times512$. Given $K$ observed sequences, we predict the next $F$ sequences. Learning this mapping in observation space is expensive, and it leaves appearance, geometry, and motion in three different formats that a predictor would have to relate on its own. \textbf{DriveUniTok} therefore maps each sequence to a single unified latent $z\in\mathbb{R}^{S\times d}$ of $S=584$ tokens with width $d=16$, from which all three modalities are decodable. \textbf{DriveUniWAM} then models the evolution of $z$ across sequences, and the DriveUniTok decoder maps predicted latents back to $I$, $D$, and $p$. In this division of labor, DriveUniTok determines which information the latent preserves, and DriveUniWAM determines how it evolves. We train DriveUniTok first, freeze it, and post-train DriveUniWAM against the NAVSIM PDM score.

\begin{figure*}[t]
\centering
\includegraphics[width=0.9\textwidth]{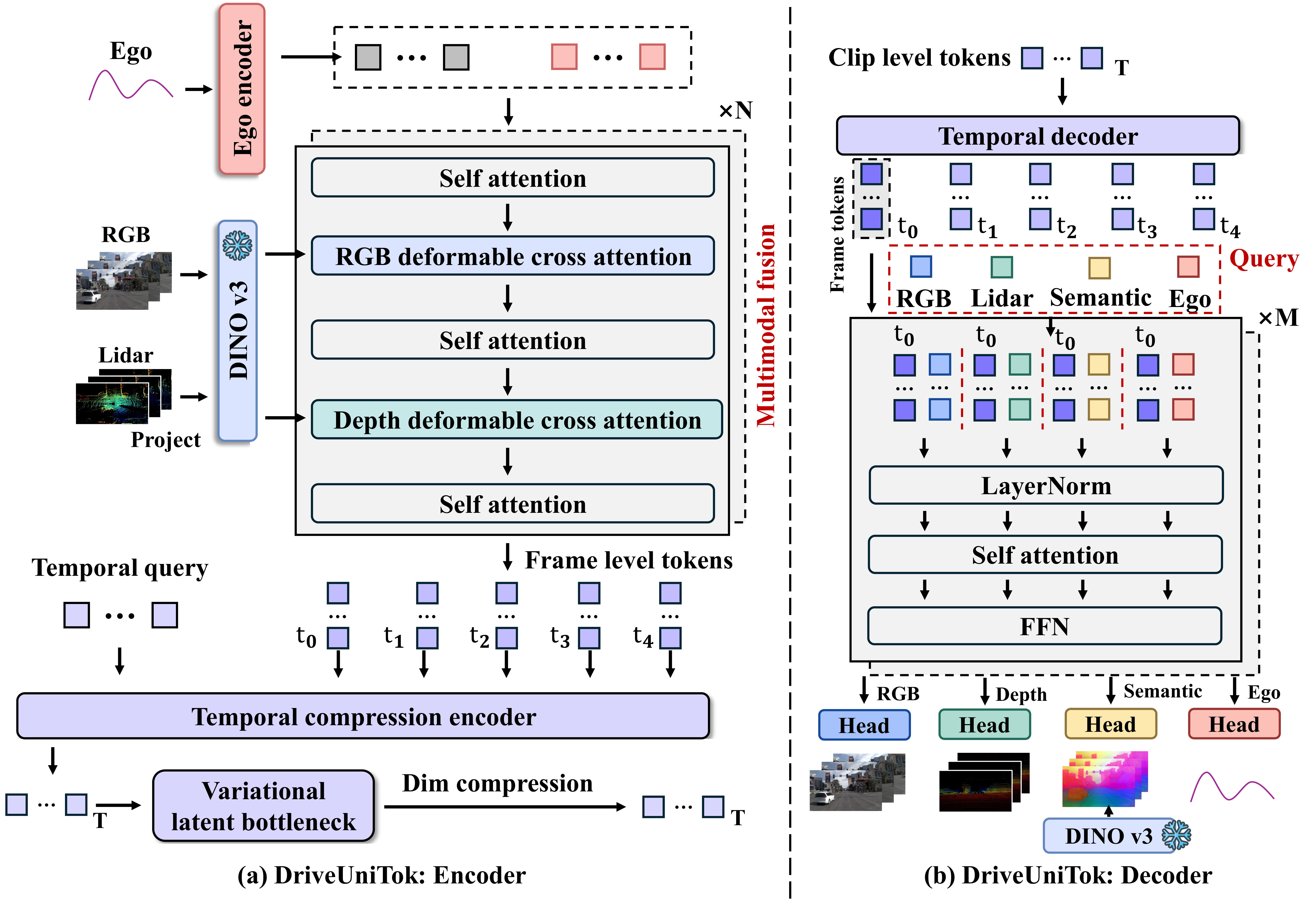}
\caption{DriveUniTok architecture. \textbf{(a) Encoder.} RGB, projected depth, and ego pose are fused into $S$ shared queries, whose $T$ per-frame states are compressed into one latent. \textbf{(b) Decoder.} The latent is expanded back to $T$ frame positions, and each target is read out through its own set of mask tokens, including a training-only semantic branch. Snowflakes mark frozen components, and all remaining modules are trained.}
\label{fig:worldtok}
\end{figure*}

\subsection{DriveUniTok: Query Aligned Multimodal Tokenizer}

DriveUniTok constructs the unified latent used by DriveUniWAM through three components: a query-based multimodal encoder that fuses the three modalities within each frame, a temporal compressor that reduces the $T$ per-frame states of a sequence to one, and a shared decoder in which one set of mask tokens per target reads all three modalities back out of the compressed latent.

\paragraph{Multimodal encoder.}
Each input stream is first converted into features. RGB frames pass through a frozen DINOv3-L backbone~\citep{simioni2025dinov3}, which yields a multilevel feature hierarchy. Rather than using a separate geometric backbone, we project the LiDAR sweep onto the image plane and process the resulting sparse depth map in the same feature space as RGB. The shared queries then extract appearance and geometric evidence through the same deformable-attention mechanism. The sparse depth map receives its own single-channel patch embedding, and the resulting tokens pass through the same frozen DINOv3 blocks with a learned modality embedding. Relative poses are mapped by a trainable ego state encoder into $N_E=8$ ego tokens. The frozen backbone therefore supplies one feature hierarchy to both RGB and depth, while the depth patch embedding and the ego state encoder only map their inputs into that shared token space. The driving-specific alignment itself is learned by the query layers described next.

The encoder uses one shared query set for all three streams. We initialize $N_I=576$ image queries on the $18\times32$ patch grid and use the ego tokens as ego queries, together forming the $S=N_I+N_E$ tokens of the latent. The two groups differ in initialization and count rather than in modality, since they are updated jointly throughout the encoder and each ends up carrying evidence from all three inputs. Writing $Q\in\mathbb{R}^{S\times C}$ for the $S$ queries of width $C$ and $F^{\mathrm{rgb}}$, $F^{\mathrm{d}}$ for the last four DINOv3 layers of the RGB and of the depth hierarchy, each encoder layer performs
\begin{equation}
Q \leftarrow \mathrm{SA}(Q),\qquad
Q \leftarrow \mathrm{SA}\big(Q + \mathrm{DA}(Q, F^{\mathrm{rgb}})\big),\qquad
Q \leftarrow \mathrm{SA}\big(Q + \mathrm{DA}(Q, F^{\mathrm{d}})\big),
\end{equation}
with $\mathrm{SA}$ self-attention over the $S$ queries and $\mathrm{DA}$ deformable cross-attention~\citep{zhu2021deformable} into a hierarchy. The same query sequence attends sequentially to RGB and depth features, while self-attention enables information exchange between the image-initialized and ego-initialized queries. This allows appearance, projected depth, and ego-pose information to interact before temporal compression. No modality is a privileged reference, and no fusion is deferred until after independent encoders have finished.

\paragraph{Temporal compression.}
Consecutive frames of a sequence are highly redundant, and keeping one latent per frame would make the downstream token sequence $T$ times longer at quadratic attention cost. Each of the $S$ queries produces one state per frame, and we compress those $T$ states into one, independently for every query. Causal self-attention first processes the $T$ states in order, then a learned pooling query gathers them into a single sequence-level state, which a Gaussian bottleneck~\citep{kingma2014vae} maps to the $d$-dimensional code that forms $z$. Because depth and ego evidence are written into these shared queries rather than into token streams of their own, a sequence yields $S$ tokens regardless of how many modalities it carries, roughly half of what separate per-modality representations would need. This gives DriveUniWAM a compact fixed-length latent in which appearance, geometry, and ego motion can be modeled together, while avoiding the duplicated token streams required by separate representations.

\paragraph{Multimodal decoder.}
A single latent step cannot be rendered directly, since the three targets are defined per frame. We therefore first project each token back to the decoder width, add learned frame queries for the $T$ temporal positions, and expand the sequence representation to $T$ frame-level memories with a temporal transformer. A shared Transformer~\citep{he2022mae,zheng2025vfmtok} then reads this memory through one set of learned mask tokens per target, each carrying its own positional embedding. The RGB and depth mask tokens yield patch features rendered by separate convolutional decoders into $\hat I$ and $\hat D$; the ego mask tokens feed the ego head that regresses $\hat p$. All three branches read the same memory, and only their mask tokens, positional embeddings, and output heads differ. This makes the multimodal alignment directly testable, since a modality absent from the shared latent could not be recovered by its branch.

Following VFMTok~\citep{zheng2025vfmtok}, we add a training-only semantic target, because pixel reconstruction alone does not preserve the semantic content of the sampled foundation model features. A fourth set of mask tokens is read by the same shared Transformer into $\hat\phi$ and aligned by cosine similarity with the final layer patch tokens $\phi(I)$ of a separate frozen DINOv3-L applied to the ground truth frames.

\paragraph{Training objective.}
DriveUniTok is trained end to end with one term per readout,
\begin{equation}
\mathcal{L}_{\mathrm{tok}}
=\lambda_{\mathrm{rgb}}\mathcal{L}_{\mathrm{rgb}}
+\lambda_{\mathrm{depth}}\mathcal{L}_{\mathrm{depth}}
+\lambda_{\mathrm{ego}}\mathcal{L}_{\mathrm{ego}}
+\lambda_{\mathrm{sem}}\mathcal{L}_{\mathrm{sem}}
+\lambda_{\mathrm{KL}}\mathcal{L}_{\mathrm{KL}}.
\end{equation}
$\mathcal{L}_{\mathrm{rgb}}$ sums $L_1$, $L_2$, VGG LPIPS~\citep{zhang2018lpips}, and ConvNeXt and DINOv3 perceptual distances. $\mathcal{L}_{\mathrm{depth}}$ is a confidence-weighted regression loss~\citep{wang2024dust3r} evaluated only where LiDAR projects to a valid pixel, plus a multiscale gradient term. $\mathcal{L}_{\mathrm{ego}}=\|\hat p-p\|_1$ and $\mathcal{L}_{\mathrm{sem}}=\mathbb{E}[1-\cos(\hat\phi,\phi(I))]$. Appendix~\ref{app:tokenizer-losses} gives every component and its weight. Sharing one decoder memory across these terms is what prevents the queries from degenerating into unrelated modality-specific subspaces while retaining the metric information driving requires.

\subsection{DriveUniWAM: Joint World-Action Modeling}

\begin{figure*}[t]
\centering
\includegraphics[width=0.9\textwidth]{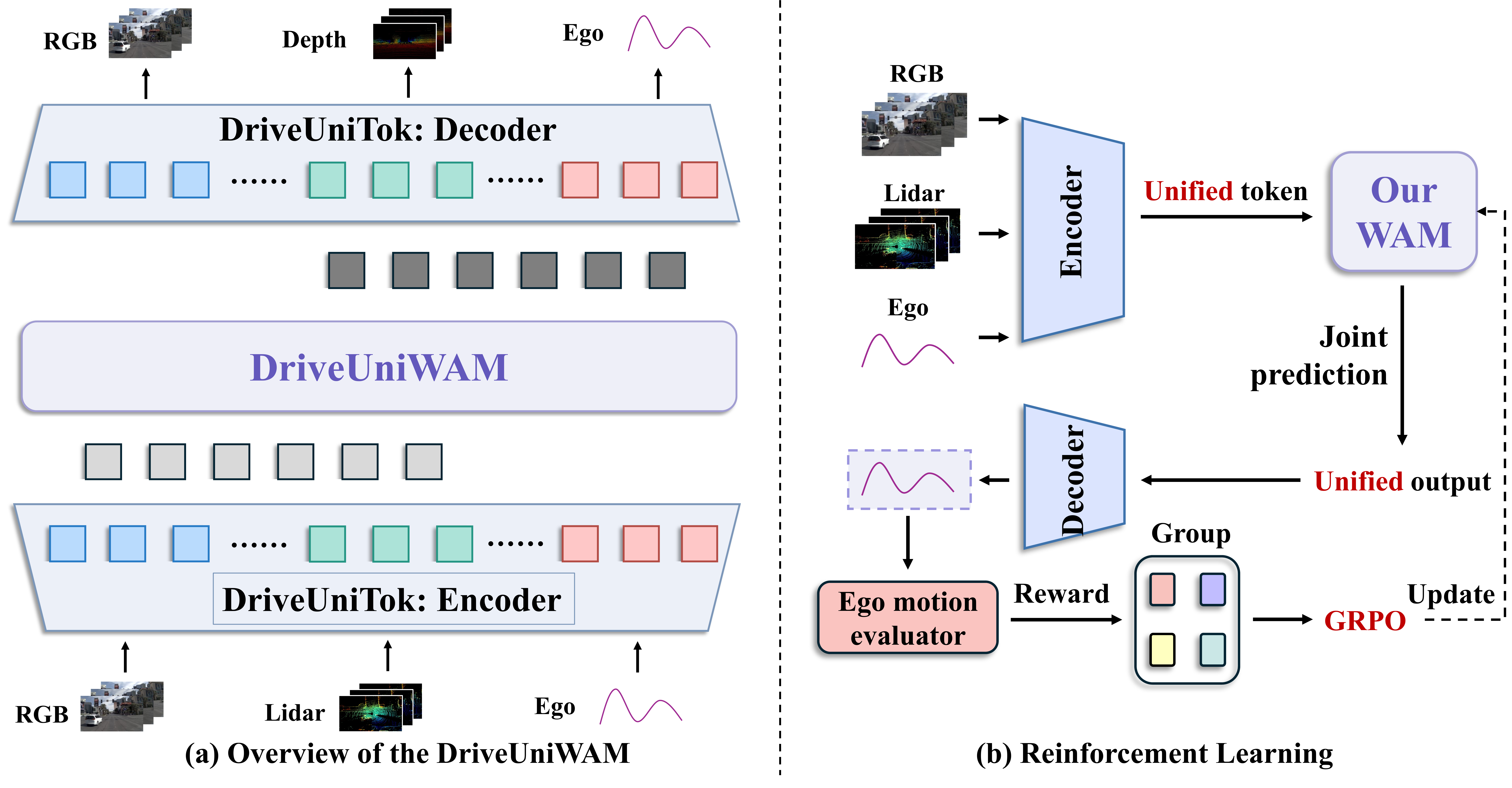}
\caption{Training and policy optimization with DriveUniWAM. \textbf{(a) Supervised training.} DriveUniWAM forecasts the future latents with flow matching. \textbf{(b) Reinforcement learning.} Per scene, a group of sampled futures is decoded into trajectories and scored by the NAVSIM PDM reward, whose relative advantages drive the Flow-GRPO update. DriveUniTok stays frozen in both stages.}
\label{fig:training}
\end{figure*}

DriveUniWAM follows the world-action model formulation introduced by DreamZero~\citep{dreamzero}. Rather than forecasting observations and leaving action generation to a separate planner, it predicts future observations and actions within one generative process, so the imagined scene evolution and the executable action share a common dynamics representation. The central role of DriveUniTok is therefore to provide a latent in which the variables needed for world-action modeling are already jointly organized. Existing world-action models commonly inherit an RGB video latent and add action tokens beside it, leaving the model to discover the correspondence between appearance and motion during training. In DriveUniWAM that correspondence is already established by the frozen DriveUniTok latent, so the denoiser learns the evolution of an aligned driving state instead of simultaneously learning two prediction tasks and their correspondence.

DriveUniWAM uses the frozen DriveUniTok to encode the $K$ observed and $F$ future sequences. The history latents provide clean context, while the future latents are corrupted and reconstructed. We initialize the denoiser from the pretrained Wan2.2-TI2V-5B model~\citep{wan2025wan} and adapt only its input and output projections to the DriveUniTok latent; a navigation embedding conditions it on the current vehicle state and route command. History queries attend only to history tokens, so the clean context cannot read noisy future targets, while every future query attends to the complete joint sequence, making all future tokens mutually visible during denoising. That visibility is what DriveUniTok makes reasonable, because each token already carries evidence from aligned appearance, geometry, and ego motion, so no artificial factorization between future observation and future action is needed.

Training uses flow matching~\citep{lipman2023flow}. A clean future latent $z_0$ is interpolated with Gaussian noise as $z_t=(1-t)z_0+t\epsilon$ at a shared time $t$, and the denoiser regresses the velocity field $v_*=\epsilon-z_0$ under
\begin{equation}
\mathcal{L}_{\mathrm{IL}}=\mathbb{E}\left[w(t)\,\|v_\theta-v_*\|_2^2\right].
\end{equation}
Because appearance, geometry, and ego motion already share the tokens of $z$, a single shared denoiser predicts the velocity field over the complete unified latent, so the model learns future scenes and ego motion as one coherent continuation, as illustrated in Figure~\ref{fig:training}(a). The unified-latent loss, latent normalization, and supervised training schedule are described in Appendix~\ref{app:sft}.

\paragraph{Reinforcement learning post-training.}
Flow matching fits recorded futures, whereas NAVSIM scores the consequences of a planned trajectory. We therefore post-train DriveUniWAM with Flow-GRPO~\citep{liu2025flowgrpo} against the NAVSIM PDM reward~\citep{navsim2024}, as shown in Figure~\ref{fig:training}(b). For each scene the policy samples a group of joint futures, the frozen decoder reads a trajectory out of each sampled latent, and the group rewards give relative advantages for a clipped policy update regularized toward the supervised reference model. Only LoRA adapters~\citep{hu2022lora} in the denoiser are trained; DriveUniTok, its decoder, and the reference policy stay frozen, so any gain reflects policy optimization around a fixed decodable latent rather than a change to the representation. Appendix~\ref{app:grpo} states the full reward, transition replay, reference BC objective, and hyperparameters.

%% file: sections/experiments.tex
\section{Experiments}

\subsection{Experimental Setup}
We conduct experiments on NAVSIM, a non-reactive planning benchmark derived from the OpenScene subset of nuPlan~\citep{navsim2024}. We train DriveUniTok on the NAVSIM training split and then freeze it before training DriveUniWAM on the same split.

We evaluate multimodal reconstruction, joint future prediction, and planning. DriveUniTok encodes front-camera RGB, projected LiDAR depth, and clip-relative ego poses into 584 tokens. Visual reconstruction uses VGG- and DINOv3-based LPIPS~\citep{zhang2018lpips,simioni2025dinov3}, PSNR, SSIM~\citep{wang2004ssim}, and I3D- and V-JEPA~2.1-based FVD~\citep{unterthiner2019fvd,carreira2017i3d,murlabadia2026vjepa21}; depth uses MAE, AbsRel, and RMSE, and ego motion uses ADE and FDE. Planning uses no-at-fault collisions (NC), drivable-area compliance (DAC), ego progress (EP), time to collision (TTC), comfort, and PDMS.

\subsection{Tokenizer Reconstruction}
\label{sec:video-tokenizer-reconstruction}

We first test whether the compact unified latent preserves useful visual and temporal information by comparing DriveUniTok with large-scale pretrained video VAEs and benchmark-scale tokenizers under a common NAVSIM navtest protocol. Baselines use official preprocessing and native temporal windows, and each source frame is evaluated at most once. Visual baselines encode RGB only, whereas DriveUniTok encodes RGB, depth, and ego pose with frozen DINOv3-L features. We report native grid compression and DriveUniTok's 584-token multimodal sequence separately to characterize fidelity--compression trade-offs under each tokenizer's native configuration. Section~\ref{sec:unified-vs-separate} further compares unified and separate tokenization under matched representation and downstream computation budgets. Results are given in Table~\ref{tab:video-tokenizer-reconstruction}; qualitative RGB and depth reconstructions are shown in Figures~\ref{fig:recon-qualitative-1}--\ref{fig:recon-qualitative-3} in Appendix~\ref{sec:qualitative-reconstructions}.

\begin{table*}[t]
\caption{Native-configuration comparison of video tokenizers on NAVSIM navtest. Comp. reports native temporal$\times$height$\times$width compression for grid tokenizers and tokens per clip for token-sequence methods; for DriveUniTok, it denotes the complete 584-token multimodal clip latent. The baselines use their official RGB-only preprocessing, whereas DriveUniTok additionally encodes projected depth and ego pose with a frozen DINOv3-L feature backbone. Results characterize fidelity--compression trade-offs under each tokenizer's native configuration.}
\label{tab:video-tokenizer-reconstruction}
\centering
\scriptsize
\setlength{\tabcolsep}{3.8pt}
\renewcommand{\arraystretch}{1.15} %
\resizebox{\textwidth}{!}{%
\begin{tabular}{l c c c c c c c c}
\toprule
& \multicolumn{2}{c}{Tokenizer configuration} & \multicolumn{6}{c}{NAVSIM navtest} \\
\cmidrule(lr){2-3} \cmidrule(lr){4-9}
Method & Comp. & Dim. & LPIPS$_{\mathrm{VGG}}\downarrow$ & LPIPS$_{\mathrm{DINO}}\downarrow$ & PSNR$\uparrow$ & SSIM$\uparrow$ & rFVD$_{\mathrm{I3D}}\downarrow$ & rFVD$_{\mathrm{VJ}}\downarrow$ \\
\midrule
\rowcolor{gray!15}
\multicolumn{9}{c}{\textit{Large-Scale Pretrained Video Tokenizers}} \\
Wan2.1 VAE~\citep{wan2025wan} & $4{\times}8{\times}8$ & 16 & 0.1151 & 0.0649 & 31.50 & 0.8819 & \textbf{2.06} & 1.0984 \\
Wan2.2 VAE~\citep{wan2025wan} & $4{\times}16{\times}16$ & 48 & \textbf{0.1081} & 0.0608 & 31.35 & 0.8775 & 2.98 & 0.7958 \\
HunyuanVideo VAE~\citep{kong2024hunyuanvideo} & $4{\times}8{\times}8$ & 16 & 0.1230 & \textbf{0.0573} & \textbf{32.36} & \textbf{0.8924} & 4.03 & \textbf{0.6278} \\
CogVideoX VAE~\citep{yang2025cogvideox} & $4{\times}8{\times}8$ & 16 & 0.1401 & 0.0943 & 32.26 & 0.8909 & 5.09 & 0.9298 \\
Cosmos-0.1 (CV4x8x8)~\citep{agarwal2025cosmos} & $4{\times}8{\times}8$ & 16 & 0.2362 & 0.3449 & 30.00 & 0.8481 & 26.76 & 4.1983 \\
Cosmos-0.1 (CV8x16x16)~\citep{agarwal2025cosmos} & $8{\times}16{\times}16$ & 16 & 0.3665 & 0.7327 & 26.03 & 0.7357 & 341.22 & 18.3995 \\
AToken~\citep{lu2026atoken} & $4{\times}16{\times}16$ & 48 & 0.1644 & 0.1105 & 29.84 & 0.8493 & 7.75 & 1.2303 \\
\rowcolor{gray!15}
\multicolumn{9}{c}{\textit{Benchmark-Scale Video Tokenizers}} \\
Open-MAGVIT2~\citep{luo2024openmagvit2} & $4{\times}8{\times}8$ & 18 & 0.1659 & 0.1308 & 25.76 & 0.7802 & 34.57 & 14.4352 \\
OmniTokenizer VAE~\citep{wang2024omnitokenizer} & $4{\times}8{\times}8$ & 8 & 0.2049 & 0.3367 & \textbf{26.89} & \textbf{0.8455} & 40.70 & 6.8585 \\
LARP-L-long~\citep{wang2025larp} & 1024 tokens & 16 & \textbf{0.1459} & 0.1332 & 26.79 & 0.8042 & \textbf{24.74} & 5.7300 \\
\textbf{DriveUniTok (Ours)} & 584 tokens & 16 & 0.1865 & \textbf{0.1063} & 25.62 & 0.7400 & 56.68 & \textbf{3.4546} \\
\bottomrule
\end{tabular}
}%
\end{table*}

DriveUniTok attains the lowest LPIPS$_{\mathrm{DINO}}$ (0.1063) and rFVD$_{\mathrm{VJ}}$ (3.4546) among benchmark-scale methods, improving over the respective second-best results by 18.7\% and 39.7\%. Large pretrained VAEs retain an advantage on pixel-oriented metrics, but do not encode geometry or ego motion. These results indicate that the compact multimodal latent preserves useful semantic and temporal information without allocating separate modality-specific token streams. We next evaluate whether this representation also benefits future prediction and planning when used by a WAM.

\subsection{Benefits of Unified Tokenization}
\label{sec:unified-vs-separate}

We next evaluate the performance and resource efficiency of unified tokenization against separate modality-specific designs for WAM prediction and planning. We compare DriveUniTok against two separate-stream controls that use the same sensors, data, DriveUniWAM backbone, and training protocol. The natural-budget control assigns a full query budget to each modality and measures the resource efficiency of the unified design. The equal-budget control uses 288 RGB, 288 depth, and 8 ego queries, matching DriveUniTok's 584 tokens, $d=16$ latent width, 9,344 latent scalars, sequence length, and downstream computation. Both models use the same frozen DINOv3-L features, semantic supervision, mask-token reconstruction objective, optimization recipe, and downstream training protocol; their architectural difference combines latent organization, parameter sharing, and cross-modal connectivity. Further implementation details are provided in Appendix~\ref{app:tokenizer-losses}. The complete comparison is reported in Table~\ref{tab:unified-vs-separate}.

\begin{table*}[t]
\caption{Unified vs.\ separate tokenization on NAVSIM. Scalars counts latent values per clip; Len.\ is the DriveUniWAM sequence length at $K{=}1,F{=}2$. Costs are in TFLOPs; DiT cost is per denoising step. Lower reconstruction and prediction errors and higher PDMS are better.}
\label{tab:unified-vs-separate}
\centering
\scriptsize
\setlength{\tabcolsep}{2.5pt}
\resizebox{\linewidth}{!}{%
\begin{tabular}{@{}l r r r r r r c c c c c c c@{}}
\toprule
& \multicolumn{3}{c}{Representation budget} & \multicolumn{3}{c}{Efficiency} & \multicolumn{3}{c}{Reconstruction} & \multicolumn{3}{c}{Future prediction} & Planning \\
\cmidrule(lr){2-4} \cmidrule(lr){5-7} \cmidrule(lr){8-10} \cmidrule(lr){11-13} \cmidrule(l){14-14}
Representation & Tokens & Scalars & Len. & \makecell{Tok.\\params (M)} & \makecell{Tok.\\cost} & \makecell{DiT/step\\cost} & LPIPS$_{\mathrm{DINO}}$ & AbsRel & ADE & gFVD$_{\mathrm{VJ}}$ & AbsRel & ADE & PDMS \\
\midrule
Separate, natural budget & 1{,}160 & 18{,}560 & 3{,}480 & 957.8 & 4.6 & 34.2 & 0.1083 & 0.0723 & \textbf{0.0702} & 7.92 & 0.1194 & 0.3997 & 90.1 \\
Separate, equal budget & 584 & 9{,}344 & 1{,}752 & 956.6 & 4.3 & 16.3 & 0.1124 & 0.0766 & 0.0720 & 12.80 & 0.1850 & 0.4186 & 87.5 \\
\textbf{Unified (DriveUniTok)} & 584 & 9{,}344 & 1{,}752 & 847.3 & 4.6 & 16.3 & \textbf{0.1063} & \textbf{0.0461} & 0.0728 & \textbf{6.61} & \textbf{0.1059} & \textbf{0.3816} & \textbf{91.0} \\
\bottomrule
\end{tabular}%
}
\end{table*}

Compared with the natural-budget control, DriveUniTok uses roughly half as many tokens and latent scalars, halves the sequence length, and reduces per-step DiT computation by about 52\%, while improving PDMS from 90.1 to 91.0. At equal representation and DiT budgets, it uses fewer tokenizer parameters, reduces gFVD$_{\mathrm{VJ}}$ by about 48\%, future depth AbsRel by about 43\%, and trajectory ADE by nearly 9\%, and raises PDMS by 3.5 points to 91.0. These results support the unified design under both budget settings. Appendices~\ref{sec:training-objective-ablation} and~\ref{sec:full-multimodal-ablation-tables} provide objective and input-modality ablations.

\subsection{Planning Performance}
Table~\ref{tab:planning} compares DriveUniWAM with end-to-end, vision-language-action, and world-action planners on NAVSIM navtest under their reported sensor configurations. DriveUniTok remains frozen during both imitation learning and reward post-training.

\begin{table*}[t]
\caption{NAVSIM navtest performance. C and L denote camera and LiDAR. The best and second-best learned results, including ties, are shown in \textbf{bold} and \underline{underlined}.}
\label{tab:planning}
\centering
\scriptsize
\setlength{\tabcolsep}{5.5pt}
\renewcommand{\arraystretch}{1.15} %
\begin{tabular}{l l l c c c c c c}
\toprule
Method & Reference & Sensors & NC$\uparrow$ & DAC$\uparrow$ & EP$\uparrow$ & TTC$\uparrow$ & Comfort$\uparrow$ & PDMS$\uparrow$ \\
\midrule
\rowcolor{gray!15}
\multicolumn{9}{c}{\textit{Reference}} \\
Human Agent & -- & -- & 100.0 & 100.0 & 87.5 & 100.0 & 99.9 & 94.8 \\
\rowcolor{gray!15}
\multicolumn{9}{c}{\textit{Vision-Based End-to-End Methods}} \\
UniAD~\citep{hu2023uniad} & CVPR'23 & 6$\times$C & 97.8 & 91.9 & 78.8 & 92.9 & \textbf{100.0} & 83.4 \\
TransFuser~\citep{chitta2022transfuser} & TPAMI'23 & 3$\times$C+L & 97.7 & 92.8 & 79.2 & 92.8 & \textbf{100.0} & 84.0 \\
ARTEMIS~\citep{feng2025artemis} & RA-L'25 & 3$\times$C+L & 98.3 & 95.1 & 81.4 & 94.3 & \textbf{100.0} & 87.0 \\
WorldRFT~\citep{yang2026worldrft} & AAAI'26 & 3$\times$C & 97.8 & 96.8 & 81.7 & 94.0 & \textbf{100.0} & 87.8 \\
DiffusionDrive~\citep{liao2025diffusiondrive} & CVPR'25 & 3$\times$C+L & 98.2 & 96.2 & 82.2 & 94.7 & \textbf{100.0} & 88.1 \\
WoTE~\citep{li2025wote} & ICCV'25 & 3$\times$C+L & 98.5 & 96.8 & 81.9 & 94.9 & \underline{99.9} & 88.3 \\
SeerDrive~\citep{zhang2025seerdrive} & NeurIPS'25 & 3$\times$C+L & 98.4 & 97.0 & 83.2 & 94.9 & \underline{99.9} & 88.9 \\
\rowcolor{gray!15}
\multicolumn{9}{c}{\textit{Vision-Language-Action Methods}} \\
UniWorldVLA~\citep{liu2026uniworldvla} & arXiv'26 & 1$\times$C & 98.7 & 96.7 & 83.2 & 96.1 & \textbf{100.0} & 89.4 \\
DriveDreamer-Policy~\citep{zhou2026drivedreamerpolicy} & arXiv'26 & 3$\times$C & 98.4 & 97.1 & 83.5 & 95.1 & \textbf{100.0} & 89.2 \\
Vega~\citep{zuo2026vega} & arXiv'26 & 1$\times$C & \underline{98.9} & 95.3 & 81.6 & 96.1 & \textbf{100.0} & 87.9 \\
ImagiDrive~\citep{li2026imagidrive} & ICRA'26 & 1$\times$C & 98.6 & 96.2 & 80.5 & 94.5 & \textbf{100.0} & 87.4 \\
AutoVLA~\citep{zhou2025autovla} & NeurIPS'25 & 3$\times$C & 98.4 & 95.6 & 81.9 & \textbf{98.0} & \underline{99.9} & 89.1 \\
ReCogDrive~\citep{li2026recogdrive} & ICLR'26 & 1$\times$C & 97.9 & 97.3 & \underline{87.3} & 94.9 & \textbf{100.0} & 90.8 \\
ExploreVLA~\citep{sheng2026explorevla} & ECCV'26 & 1$\times$C & 98.8 & \underline{98.4} & 83.5 & 96.5 & \underline{99.9} & 90.4 \\
DriveVLA-W0~\citep{li2026drivevlaw0} & ICLR'26 & 1$\times$C & 98.7 & \textbf{99.1} & 83.3 & 95.3 & 99.3 & 90.2 \\
SGDrive~\citep{li2026sgdrive} & CVPR'26 & 1$\times$C & 98.6 & 97.8 & 85.8 & 96.2 & \textbf{100.0} & \underline{91.1} \\
\rowcolor{gray!15}
\multicolumn{9}{c}{\textit{World-Action Models}} \\
Epona~\citep{zhang2025epona} & ICCV'25 & 1$\times$C & 97.9 & 95.1 & 80.4 & 93.8 & \underline{99.9} & 86.2 \\
PWM~\citep{zhao2025pwm} & NeurIPS'25 & 1$\times$C & 98.6 & 95.9 & 81.8 & 95.4 & \textbf{100.0} & 88.1 \\
DriveLaW~\citep{xia2026drivelaw} & CVPR'26 & 1$\times$C & \textbf{99.0} & 97.1 & 81.3 & 96.7 & \textbf{100.0} & 89.1 \\
DriveWAM~\citep{shi2026drivewam} & arXiv'26 & 1$\times$C & 98.3 & 98.1 & 84.3 & 95.2 & \textbf{100.0} & 90.1 \\
\textbf{DriveUniWAM (Camera-only)} & -- & 1$\times$C & \underline{98.9} & 97.7 & 84.0 & \underline{96.8} & \underline{99.9} & 90.5 \\
\textbf{DriveUniWAM (Ours)} & -- & 1$\times$C+L & \textbf{99.0} & 97.7 & 85.6 & 96.3 & \textbf{100.0} & 91.0 \\
\textbf{DriveUniWAM + RL (Ours)} & -- & 1$\times$C+L & 98.7 & 98.0 & \textbf{87.4} & 95.6 & 99.8 & \textbf{91.54} \\
\bottomrule
\end{tabular}
\end{table*}

DriveUniWAM reaches 90.5 PDMS with camera-only input, exceeding DriveWAM by 0.4 points. Adding projected LiDAR raises PDMS to 91.0, a further gain of 0.5 points and a 0.9-point improvement over DriveWAM. Reward post-training further increases PDMS to 91.54. With the tokenizer fixed, this gain reflects policy optimization within the same decodable latent space. DriveUniWAM also obtains 90.3 EPDMS on NAVSIM-v2; additional results and qualitative trajectories are provided in Appendix~\ref{sec:navsim-v2-results} and Figures~\ref{fig:navtest-qualitative-1}--\ref{fig:navtest-qualitative-10} in Appendix~\ref{sec:qualitative-navsim}.

\subsection{Backbone and Tokenizer Generalization}
Table~\ref{tab:backbone-flexibility} tests DriveUniTok across video diffusion backbones. Part I replaces the native VAE while keeping the backbone fixed; Part II compares different backbones using DriveUniTok. Replacing the native VAE improves PDMS by 2.5 points for Wan2.1-1.3B and 3.1 points for Wan2.2-5B. Across five backbones, the unified latent achieves 88.9--91.0 PDMS, supporting its utility across model families and scales.

\begin{table*}[t]
\caption{Ablation of video backbones and latent representations on the NAVSIM navtest benchmark. Part I compares the native VAE and DriveUniTok with fixed backbones, while Part II evaluates DriveUniWAM across video DiT backbones using the same unified latent. Across both parts, best and second-best results are shown in \textbf{bold} and \underline{underlined}; improvements over the native VAE appear in parentheses.}
\label{tab:backbone-flexibility}
\centering
\footnotesize
\setlength{\tabcolsep}{6.5pt}
\renewcommand{\arraystretch}{1.15}
\resizebox{0.92\textwidth}{!}{%
\begin{tabular}{l c c c c c l}
\toprule
Video backbone & Tokenizer & NC$\uparrow$ & DAC$\uparrow$ & EP$\uparrow$ & TTC$\uparrow$ & PDMS$\uparrow$ \\
\midrule
\rowcolor{gray!15}
\multicolumn{7}{c}{\textit{Part I: Visual Tokenizer Representation Ablation}} \\
\multirow{2}{*}{Wan2.1-T2V-1.3B \citep{wan2025wan}} & Native VAE & 98.1 & 94.7 & 80.9 & 94.5 & 86.5 \\
& DriveUniTok & 98.4 & \underline{97.0} & 82.9 & 95.4 & 89.0 \scriptsize{(+2.5)} \\
\addlinespace[4pt]
\multirow{2}{*}{Wan2.2-TI2V-5B \citep{wan2025wan}} & Native VAE & 98.4 & 95.5 & 82.6 & 94.8 & 87.9 \\
& DriveUniTok & \underline{99.0} & \textbf{97.7} & \textbf{85.6} & \underline{96.3} & \textbf{91.0} \scriptsize{\textbf{(+3.1)}} \\
\midrule
\rowcolor{gray!15}
\multicolumn{7}{c}{\textit{Part II: Scalability across Video DiT Backbones (with DriveUniTok)}} \\
Wan2.1-T2V-1.3B \citep{wan2025wan} & DriveUniTok & 98.4 & \underline{97.0} & 82.9 & 95.4 & 89.0 \\
CogVideoX-2B \citep{yang2025cogvideox} & DriveUniTok & 98.9 & 96.5 & 82.5 & 96.2 & 88.9 \\
Cosmos-Predict2.5-2B \citep{agarwal2025cosmos} & DriveUniTok & \underline{99.0} & 96.2 & 83.3 & 96.1 & 89.1 \\
Wan2.2-TI2V-5B \citep{wan2025wan} & DriveUniTok & \underline{99.0} & \textbf{97.7} & \textbf{85.6} & \underline{96.3} & \textbf{91.0} \\
HunyuanVideo-1.5-8.3B \citep{hunyuanvideo2025} & DriveUniTok & \textbf{99.1} & \textbf{97.7} & \underline{84.0} & \textbf{96.8} & \underline{90.5} \\
\bottomrule
\end{tabular}
}%
\end{table*}

%% file: sections/conclusion.tex
\section{Conclusion}
\label{sec:conclusion}

We presented DriveUniTok and DriveUniWAM for joint future-scene and ego-trajectory prediction through a unified multimodal latent. On NAVSIM, the unified design improves prediction and planning with fewer tokenizer parameters and lower downstream computational cost than the natural-budget separate-stream control. DriveUniWAM achieves 91.0 PDMS with supervised training and 91.54 after reinforcement learning post-training.

%% file: sections/ai_use_statement.tex
\section*{AI Use Statement}

We used AI tools to assist with code development and debugging, manuscript language editing, and literature search and review. The tools were used as research assistance rather than as a substitute for author verification of technical claims. The authors checked the implementation, experimental results, citations, and final manuscript, and take full responsibility for the content of this paper.

%% file: sections/reproducibility_statement.tex
\section*{Reproducibility Statement}

To support reproducibility, the main paper and Appendices~\ref{app:tokenizer-losses},~\ref{app:sft}, and~\ref{app:grpo} document the model, objectives, training procedures, and evaluation details needed to reproduce our DriveUniTok tokenizer, DriveUniWAM training, and NAVSIM evaluation.

%% file: sections/supplementary.tex
\appendix
\section*{Appendix}
\section{Tokenizer Implementation Details}
\label{app:tokenizer-losses}

\paragraph{Encoder and decoder configuration.}
Each input is resized to $288\times512$ and divided into patches of size $16$, yielding an $18\times32$ grid. The encoder is initialized with $N_I=576$ queries with spatial reference points and $N_E=8$ queries with learned reference points, all with embedding width $1024$. These query groups differ in initialization rather than modality: after joint attention, they form one unified sequence of $S=584$ tokens, with every token able to carry evidence from RGB, depth, and ego motion. The DINOv3-L feature backbone is initialized from a pretrained checkpoint and kept frozen throughout tokenizer training; the separate DINOv3 semantic teacher is also frozen. The encoder contains two query layers; each layer applies joint self-attention over the $S$ queries followed by deformable cross-attention \citep{zhu2021deformable} over the RGB and depth feature hierarchies. We use four feature levels, four sampling points per level, and $16$ attention heads. Reference points for the first query group are fixed at patch-grid centers, whereas the second query group uses learned reference points because it has no spatial prior.

Temporal compression and expansion each use two causal self-attention blocks. The $T$ states of each unified query are temporally pooled into one sequence-level state and passed through a Variational Latent Bottleneck (VLB), which compresses the state to width $d=16$. Thus, DriveUniTok produces one unified $z\in\mathbb{R}^{S\times16}$ rather than separate image and ego latents. The decoder maps this latent back to width $1024$ and applies $24$ shared ViT blocks with $16$ heads and full self-attention. Four sets of mask tokens read the shared decoder features; the RGB and depth branches use convolutional decoders with channel multipliers $[1,1,2,2,4]$ and two residual blocks per stage. We optimize all trainable tokenizer modules with AdamW at learning rate $10^{-4}$ and weight decay $0.01$, clip gradients at norm $10$, and use warmup followed by a cosine schedule. The query-aggregator parameter group uses a $0.1\times$ learning-rate multiplier. All DINOv3 parameters receive no gradient updates. This unified latent is the representation consumed by DriveUniWAM for joint future prediction and planning.

\paragraph{Equal-budget unified--separate comparison.}
\label{app:equal-budget-details}
We provide additional implementation details for the comparison in
Table~\ref{tab:unified-vs-separate}. The protocol matches the representation exposed to DriveUniWAM, allowing
us to compare the downstream effectiveness of the unified and separate
designs under matched representation and downstream computation budgets.
DriveUniTok uses 584 unified tokens, initialized from 576 queries with
spatial reference points and 8 queries with learned reference points. The equal-budget
separate baseline uses 288 RGB queries, 288 projected-LiDAR depth
queries, and 8 ego queries, also totaling 584 tokens. Both RGB and
depth query sets use $12\times24$ grids. With latent width $d=16$,
both models therefore produce $584\times16=9{,}344$ latent scalars and
the same downstream sequence length of $3\times584=1{,}752$ tokens,
giving the same nominal self-attention budget.

Both arms use the same frozen DINOv3-L feature extractor, frozen
semantic teacher, RGB/depth reconstruction objectives, semantic
auxiliary loss, depth preprocessing, and mask-token reconstruction
path. The semantic supervision configuration and loss weights are kept
unchanged. Neither arm uses a separate masking stage; mask-token reconstruction is
part of the same tokenizer training procedure for both models.

The architectural difference is the location and connectivity of
cross-modal interaction. DriveUniTok uses one shared query sequence, allowing
RGB, depth, and ego queries to exchange information through self-
attention before temporal compression and the Variational Latent
Bottleneck. The separate baseline uses independent RGB, depth, and ego
streams: RGB queries attend only to RGB features, depth queries attend
only to depth features, and ego queries do not attend to visual
features. Each stream has its own query set, encoder stack, temporal
compressor, and Variational Latent Bottleneck. The separate design
therefore removes cross-modal interaction in the tokenizer rather than
merely assigning different output slices to a shared encoder.

The observation decoders use the same ViT-based decoder configuration,
including the same depth, number of attention heads, and MLP ratio of
$4.0$. For
the separate baseline, decoder context is restricted to the
corresponding token ranges: RGB tokens occupy $[0,288)$, depth tokens
occupy $[288,576)$, and ego tokens occupy $[576,584)$. The unified model
uses the full shared token sequence as decoder context. In the reported
separate-stream configuration, ego reconstruction directly uses the
aggregated ego tokens rather than the shared decoder ego readout.

We match the tokenizer training recipe across the two arms, including
the data protocol, batch size, learning rate of $10^{-4}$, precision,
gradient accumulation, learning-rate schedule, and 33K tokenizer
updates. We likewise match the downstream DriveUniWAM recipe, including the
same data protocol, model configuration, 130K updates, learning rate,
warm-up schedule, and flow-shift setting.

The equal-budget condition matches token count, latent-scalar count,
downstream sequence length, nominal attention cost, sensors, training
data, and optimization recipe. It does not imply exact equality of
trainable parameter counts: the separate baseline contains independent
copies of the query sets, encoder stacks, temporal compressors, and
Variational Latent Bottlenecks, whereas DriveUniTok uses a shared aggregation
pathway. The frozen DINOv3-L backbone and shared depth patch embedding
are common to both models and are not treated as additional trainable
capacity. The equal-budget condition therefore matches the representation
and downstream DiT budgets while exposing the parameter-efficiency
difference between the two tokenizer designs. Together with the
natural-budget comparison, this evaluates whether the unified design
remains advantageous under both lower-resource and matched-budget settings.

\paragraph{Loss components and weights.}
We use $\lambda_{\mathrm{rgb}}=\lambda_{\mathrm{depth}}=\lambda_{\mathrm{ego}}=\lambda_{\mathrm{sem}}=1$ and $\lambda_{\mathrm{KL}}=0.1$. Table~\ref{tab:tokenizer-loss-weights} expands $\mathcal{L}_{\mathrm{tok}}$ into all of its component terms and lists the weight assigned to each term within its group.

\begin{table}[h]
\caption{Components of the DriveUniTok training objective. $\Omega$ denotes the set of pixels where LiDAR projects to a valid depth.}
\label{tab:tokenizer-loss-weights}
\centering
\small
\begin{tabular}{l l l c}
\toprule
Group & Term & Form & Weight \\
\midrule
\multirow{5}{*}{$\mathcal{L}_{\mathrm{rgb}}$}
 & pixel $L_1$        & $\|\hat I-I\|_1$                                   & 1.0 \\
 & pixel $L_2$        & $\|\hat I-I\|_2^2$                                 & 1.0 \\
 & LPIPS              & VGG feature distance \citep{zhang2018lpips}         & 1.0 \\
 & perceptual         & ConvNeXt feature distance                          & 0.1 \\
 & perceptual         & DINOv3 feature distance                            & 1.0 \\
\midrule
\multirow{3}{*}{$\mathcal{L}_{\mathrm{depth}}$}
 & confidence         & $\mathbb{E}_{\Omega}[r\kappa-\alpha\log\kappa]$, $\alpha{=}0.2$ & 1.0 \\
 & regression         & $\mathbb{E}_{\Omega}[r]$, $r=|\hat D-D|$           & 10.0 \\
 & gradient           & multiscale gradient matching on $\Omega$           & 0.1 \\
\midrule
$\mathcal{L}_{\mathrm{ego}}$ & pose $L_1$ & $\|\hat p-p\|_1$ on $(x,y,\text{heading})$ & 1.0 \\
$\mathcal{L}_{\mathrm{sem}}$ & cosine     & $\mathbb{E}[1-\cos(\hat\phi,\phi(I))]$      & 1.0 \\
\midrule
$\mathcal{L}_{\mathrm{KL}}$ & Gaussian KL & unified VLB & $0.1$ \\
\bottomrule
\end{tabular}
\end{table}

For valid projected pixels $\Omega$, the depth confidence term uses $r\kappa-\alpha\log\kappa$ with $r=\lVert\hat D-D\rVert_2$, $\alpha=0.2$, and $\kappa$ the predicted confidence. The regression term uses the same valid pixels, and both confidence and regression values are filtered at the $0.98$ quantile. Depth is clipped to $[1,150]$\,m and normalized by an average scale of $60$. Batches with fewer than 100 valid points skip the depth terms. The semantic target is the final-layer patch tokens of a frozen DINOv3-L applied to the ground-truth frames, which is a separate instance from the frozen encoder backbone.

\FloatBarrier
\section{Qualitative Tokenizer Reconstructions}
\label{sec:qualitative-reconstructions}

Each group compares the ground-truth RGB image, its reconstruction from the unified latent, the LiDAR-projected ground-truth depth, and the reconstructed depth at valid LiDAR pixels. Across nine scenes and five consecutive frames per scene, the RGB reconstructions preserve road geometry, vehicles, and salient scene details, while the depth reconstructions retain the spatial structure and distance patterns present in the sparse LiDAR measurements. The comparisons provide a qualitative complement to the reconstruction metrics in Table~\ref{tab:video-tokenizer-reconstruction}: the tokenizer preserves both appearance and metric 3D structure within the same compact representation.

\begin{figure*}[tb]
\centering
\includegraphics[height=0.82\textheight,keepaspectratio]{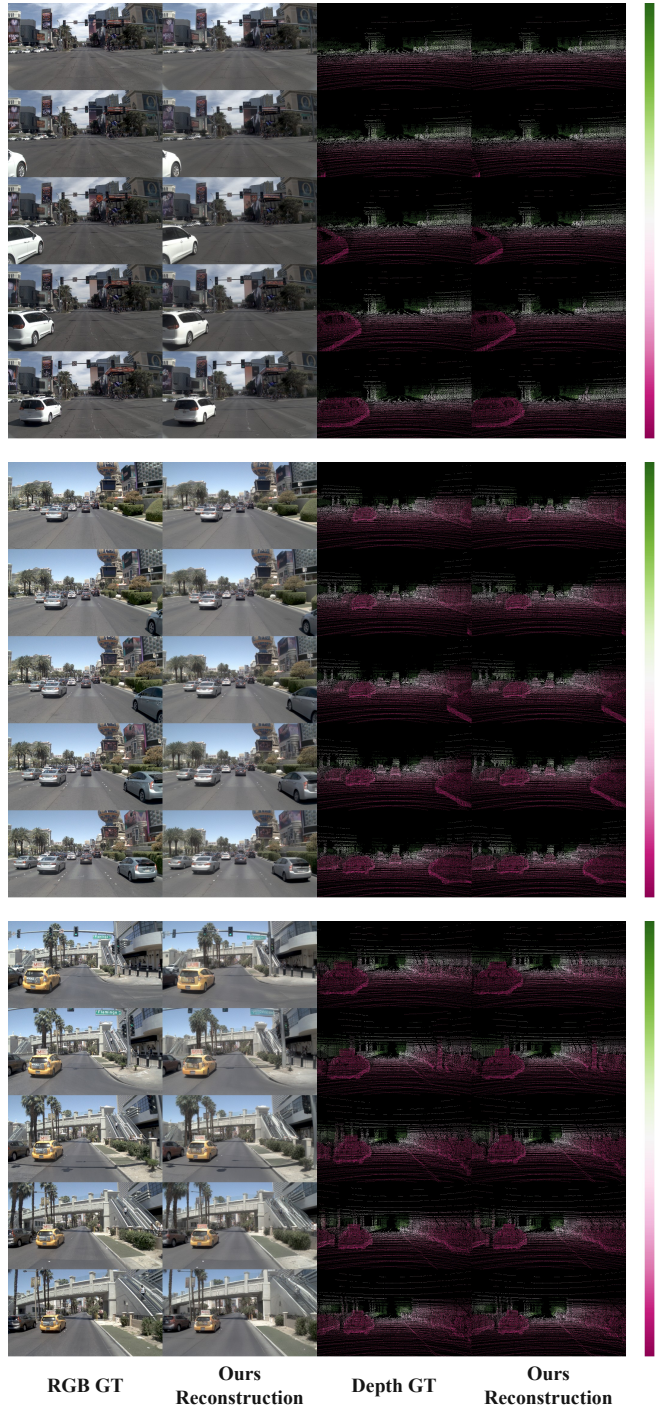}
\caption{Qualitative tokenizer reconstructions for reconstruction group 1. From left to right, the columns show RGB ground truth, reconstructed RGB, LiDAR-projected depth ground truth, and reconstructed depth at valid LiDAR pixels.}
\label{fig:recon-qualitative-1}
\end{figure*}

\begin{figure*}[tb]
\centering
\includegraphics[height=0.82\textheight,keepaspectratio]{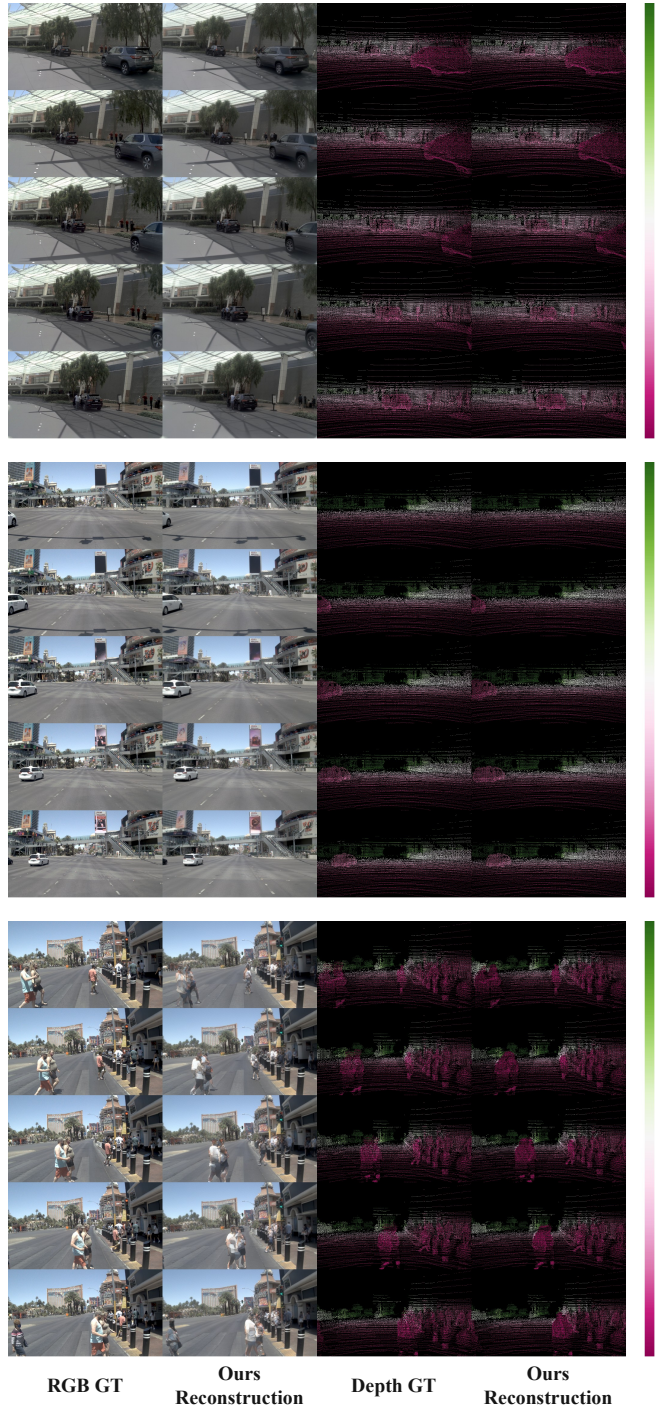}
\caption{Qualitative tokenizer reconstructions for reconstruction group 2. From left to right, the columns show RGB ground truth, reconstructed RGB, LiDAR-projected depth ground truth, and reconstructed depth at valid LiDAR pixels.}
\label{fig:recon-qualitative-2}
\end{figure*}

\begin{figure*}[tb]
\centering
\includegraphics[height=0.82\textheight,keepaspectratio]{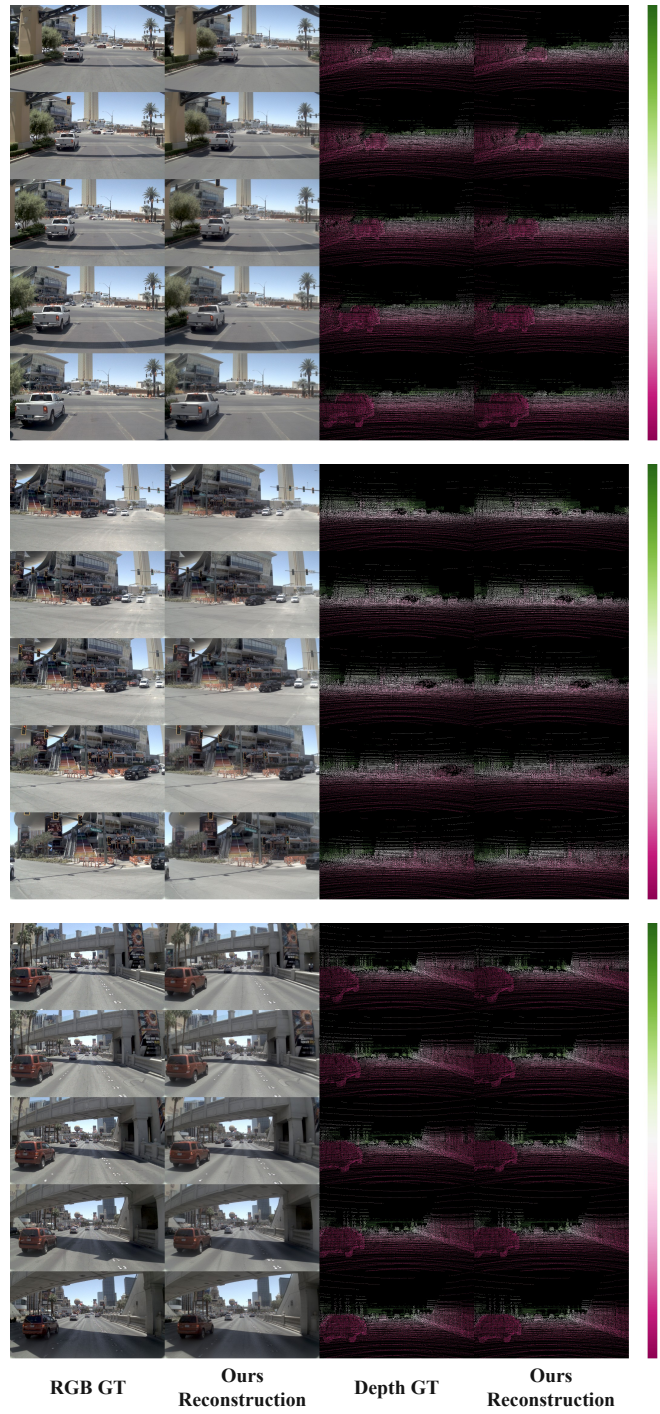}
\caption{Qualitative tokenizer reconstructions for reconstruction group 3. From left to right, the columns show RGB ground truth, reconstructed RGB, LiDAR-projected depth ground truth, and reconstructed depth at valid LiDAR pixels.}
\label{fig:recon-qualitative-3}
\end{figure*}

\FloatBarrier

\section{DriveUniWAM Training Details}
\label{app:wam-training}

\subsection{Supervised Flow Matching}
\label{app:sft}

The supervised model uses one history clip and two future clips, corresponding to the $K{=}1,F{=}2$ setting. Each clip contains five frames sampled at 2\,Hz. With shared boundary frames, the history and future clips cover frame indices $[0,1,2,3,4]$, $[4,5,6,7,8]$, and $[8,9,10,11,12]$, respectively. Each clip's ego pose is expressed relative to its first frame. When clips are concatenated, the poses of each subsequent clip are transformed into the coordinate frame anchored at the endpoint of the preceding clip, so the shared boundary frames produce a continuous trajectory. The unified latent is normalized per channel using statistics estimated on the training split. We sample Gaussian noise $\epsilon$ and a shifted flow time $\sigma$ with shift $3.0$, then construct
\begin{equation}
z_{\sigma}=(1-\sigma)z_{0}+\sigma\epsilon,
\qquad
v_*=\epsilon-z_{0}.
\end{equation}
The history clip remains clean, while the two future clips share the same $\sigma$ and are noised independently. A single shared denoiser predicts the velocity field over the complete unified latent:
\begin{equation}
\mathcal{L}_{\mathrm{SFT}}
=\mathbb{E}\left[w(\sigma)\left\|v_{\theta}-v_*\right\|_2^2\right].
\end{equation}
The loss is averaged over the complete unified token sequence. The navigation condition has eight values: four current-frame kinematic values and four route-command values. Only the Wan DiT parameters are optimized; DriveUniTok and its decoder remain frozen. The reported training uses AdamW with learning rate $10^{-4}$, betas $(0.9,0.95)$, weight decay $0.01$, 1,000 warmup updates, cosine decay, gradient clipping at $1.0$, and 130,000 optimizer updates.

\subsection{Reinforcement Learning Post-Training}
\label{app:grpo}

We post-train DriveUniWAM with Flow-GRPO \citep{liu2025flowgrpo} against the NAVSIM PDM reward \citep{navsim2024}. For each scene the policy samples a group of $G=8$ joint futures over $10$ stochastic Flow-SDE denoising steps, and the frozen DriveUniTok decoder reads a trajectory out of each sampled latent for NAVSIM to score. For rewards $\{R_i\}_{i=1}^{G}$ the group relative advantage is
\begin{equation}
A_i=\frac{R_i-\operatorname{mean}_{j}(R_j)}
{\operatorname{std}_{j}(R_j)+\varepsilon}.
\end{equation}
Let $x_{i,k-1}$ denote the sampled next state from $x_{i,k}$ in the denoising chain. With the transition likelihood ratio $r_{i,k}(\theta)=\exp[\log p_{\theta}(x_{i,k-1}\mid x_{i,k})-\log p_{\mathrm{old}}(x_{i,k-1}\mid x_{i,k})]$ we optimize the clipped objective
\begin{equation}
\mathcal{L}_{\mathrm{policy}}
=-\mathbb{E}_{i,k}\left[\gamma^{N-k-1}
\min\left(r_{i,k}(\theta)A_i,
\operatorname{clip}(r_{i,k}(\theta),1-\delta,1+\delta)A_i\right)\right],
\end{equation}
where $N$ is the number of denoising steps, $\gamma=0.6$ weights transitions by their contribution to the final sample, and $\delta=0.02$. Since the trajectory is decoded from the unified latent, the likelihood ratio is computed over the transitions of the complete token sequence. A frozen reference model supplies a behavior cloning regularizer over the unified latent transition mean:
\begin{equation}
\mathcal{L}_{\mathrm{ref}}
=\frac{\|\mu_\theta^{z}-\mu_{\mathrm{ref}}^{z}\|_2^2}{2\sigma^2},
\end{equation}
This regularizer keeps the policy close to the reference model on the decodable unified-latent manifold. The final objective is $\mathcal{L}_{\mathrm{RL}}=\mathcal{L}_{\mathrm{policy}}+\lambda_{\mathrm{ref}}\mathcal{L}_{\mathrm{ref}}$, with $\lambda_{\mathrm{ref}}=0.1$. We train rank-$16$, scale-$32$ LoRA adapters \citep{hu2022lora} on the attention projections for four inner epochs per rollout at learning rate $5\times10^{-5}$. DriveUniTok, its decoder, and the reference policy remain frozen throughout post-training.

\section{Training Objective Ablation}
\label{sec:training-objective-ablation}

This ablation isolates the masking and semantic-supervision objectives from the representation comparison. During tokenizer training on NAVSIM, intra-modal and cross-modal masks are sampled randomly for each training example. Intra-modal masking removes patches or tokens within an observed modality and reconstructs them from the remaining evidence in that modality. Cross-modal masking hides one target modality and reconstructs it from the others. Semantic supervision refers to the training-only semantic feature reconstruction branch; when it is removed, the RGB DINO perceptual objective is retained, so the ablation affects only the semantic readout. All variants use the same unified architecture, latent budget, NAVSIM training split, optimization budget, and full-input NAVSIM navtest evaluation protocol. The results are summarized in Table~\ref{tab:training-objective-ablation}.

\begin{table*}[t]
\caption{Training-objective ablation on NAVSIM navtest. Each row removes the named objectives from the full recipe. All variants share the architecture, $584\times16$ latent, and complete camera, depth, and ego inputs at evaluation. Best values, including ties, are bold.}
\label{tab:training-objective-ablation}
\centering
\scriptsize
\setlength{\tabcolsep}{2.5pt}
\begin{tabular}{@{}l c c c c c c c c@{}}
\toprule
& \multicolumn{4}{c}{RGB} & \multicolumn{2}{c}{Depth} & \multicolumn{2}{c}{Ego} \\
\cmidrule(lr){2-5} \cmidrule(lr){6-7} \cmidrule(l){8-9}
Variant & LPIPS$_{\mathrm{DINO}}\downarrow$ & PSNR$\uparrow$ & SSIM$\uparrow$ & rFVD$_{\mathrm{VJ}}\downarrow$ & AbsRel$\downarrow$ & RMSE$\downarrow$ & ADE$\downarrow$ & FDE$\downarrow$ \\
\midrule
\makecell[l]{w/o masked reconstruction\\and semantic supervision} &
0.1540 & 24.55 & 0.68 & 8.40 & 0.0695 & 9.30 & 0.1450 & 0.2670 \\
w/o intra-modal masking & 0.1215 & 25.20 & 0.72 & 4.85 & 0.0520 & 8.00 & 0.0900 & 0.1650 \\
w/o cross-modal masking & 0.1298 & 25.02 & 0.72 & 5.60 & 0.0565 & 8.30 & 0.1040 & 0.1910 \\
w/o semantic supervision & 0.1195 & 25.55 & \textbf{0.74} & 4.28 & 0.0480 & 7.72 & 0.0775 & 0.1420 \\
\textbf{Full training objective} & \textbf{0.1063} & \textbf{25.62} & \textbf{0.74} & \textbf{3.45} & \textbf{0.0461} & \textbf{7.57} & \textbf{0.0728} & \textbf{0.1329} \\
\bottomrule
\end{tabular}
\end{table*}

This ablation complements the equal-budget comparison by examining how the training objectives affect the quality of the unified latent itself. The full recipe achieves the best results across all reported RGB, depth, and ego metrics. Compared with omitting masked reconstruction and semantic supervision, it reduces LPIPS$_{\mathrm{DINO}}$ from 0.1540 to 0.1063, rFVD$_{\mathrm{VJ}}$ from 8.4000 to 3.4546, depth AbsRel from 0.0695 to 0.0461, and trajectory ADE from 0.1450 to 0.0728. These results indicate that both the cross-modal reconstruction objective and the auxiliary semantic target contribute to the quality of the unified latent used by DriveUniWAM.

\section{NAVSIM-v2 Planning Results}
\label{sec:navsim-v2-results}

We further evaluate the planning model under the NAVSIM-v2 protocol, which includes additional scene- and motion-compliance criteria. Table~\ref{tab:planning-v2} shows that DriveUniWAM reaches 90.3 EPDMS while maintaining high scores across the individual metrics. This result supports the effectiveness of the unified latent representation for WAM planning: it preserves scene evidence, semantic structure, and the metric ego state needed to evaluate future action sequences.

\begin{table*}[t]
\caption{Performance on the NAVSIM-v2 navtest benchmark. Human Agent is included as a reference. DriveUniWAM (Ours) denotes the imitation-trained DriveUniWAM with a frozen DriveUniTok tokenizer. TLC denotes traffic-light compliance, HC denotes history comfort, EC denotes extended comfort, and EPDMS is the NAVSIM-v2 aggregate score. The best and second-best learned results, including ties, are shown in \textbf{bold} and \underline{underlined}.}
\label{tab:planning-v2}
\centering
\scriptsize
\setlength{\tabcolsep}{4.2pt}
\renewcommand{\arraystretch}{1.15}
\resizebox{\textwidth}{!}{%
\begin{tabular}{l l c c c c c c c c c c}
\toprule
Method & Reference & NC$\uparrow$ & DAC$\uparrow$ & DDC$\uparrow$ & TLC$\uparrow$ & EP$\uparrow$ & TTC$\uparrow$ & LK$\uparrow$ & HC$\uparrow$ & EC$\uparrow$ & EPDMS$\uparrow$ \\
\midrule
\rowcolor{gray!15}
\multicolumn{12}{c}{\textit{Reference}} \\
Human Agent & -- & 100.0 & 100.0 & 99.8 & 100.0 & 87.4 & 100.0 & 100.0 & 98.1 & 90.1 & 94.5 \\
\rowcolor{gray!15}
\multicolumn{12}{c}{\textit{Vision-Based End-to-End Methods}} \\
TransFuser~\citep{chitta2022transfuser} & TPAMI'23 & 96.90 & 89.90 & 97.80 & 99.70 & 87.10 & 95.40 & 92.70 & \underline{98.30} & \underline{87.20} & 76.7 \\
DriveSuprim~\citep{yao2026drivesuprim} & AAAI'26 & 97.50 & 96.50 & 99.40 & 99.60 & \underline{88.40} & 96.60 & 95.50 & \underline{98.30} & 77.00 & 83.1 \\
DiffusionDrive~\citep{liao2025diffusiondrive} & CVPR'25 & 98.20 & 95.90 & 99.40 & 99.80 & 87.50 & 97.30 & 96.80 & \underline{98.30} & \textbf{87.70} & 84.5 \\
\rowcolor{gray!15}
\multicolumn{12}{c}{\textit{Vision-Language-Action Methods}} \\
ReCogDrive~\citep{li2026recogdrive} & ICLR'26 & 98.30 & 95.20 & 99.50 & 99.80 & 87.10 & 97.50 & 96.60 & \underline{98.30} & 86.50 & 83.6 \\
SGDrive~\citep{li2026sgdrive} & CVPR'26 & 98.60 & 94.30 & 99.50 & \textbf{99.90} & 86.00 & 97.90 & 96.10 & \underline{98.30} & 85.90 & 86.2 \\
DriveFine~\citep{dang2026drivefine} & arXiv'26 & \textbf{98.70} & 97.30 & 99.50 & 99.80 & \textbf{88.70} & 97.80 & 97.70 & \textbf{98.40} & 83.80 & \underline{89.7} \\
DriveVLA-W0~\citep{li2026drivevlaw0} & ICLR'26 & 98.50 & \textbf{99.10} & 98.00 & 99.70 & 86.40 & 98.10 & 93.20 & 97.90 & 58.90 & 86.1 \\
\rowcolor{gray!15}
\multicolumn{12}{c}{\textit{World-Action Models}} \\
DVGT-2~\citep{zuo2026dvgt2} & arXiv'26 & \textbf{98.70} & 97.90 & \textbf{99.70} & \textbf{99.90} & 87.90 & 98.00 & \textbf{98.20} & \textbf{98.40} & 77.00 & 89.6 \\
EponaV2~\citep{xu2026eponav2} & arXiv'26 & 98.50 & 97.40 & 99.50 & \textbf{99.90} & 87.90 & 98.10 & 97.70 & 98.20 & 77.40 & 88.9 \\
DriveLaW~\citep{xia2026drivelaw} & CVPR'26 & \textbf{98.70} & 96.90 & \underline{99.60} & 99.80 & 87.50 & \underline{98.30} & 97.60 & \textbf{98.40} & 77.40 & 88.6 \\
\textbf{DriveUniWAM (Ours)} & -- & \underline{98.63} & \underline{98.11} & \underline{99.60} & \underline{99.88} & 88.11 & \textbf{98.51} & \underline{97.83} & 98.18 & 80.65 & \textbf{90.3} \\
\bottomrule
\end{tabular}
}%
\end{table*}

\section{Modality Sensitivity}
\label{sec:full-multimodal-ablation-tables}

We finally vary the available input modalities to examine how different sources of evidence contribute to the unified latent, reconstruction, and joint future generation. Table~\ref{tab:modality-ablation} reports evaluation-time modality sensitivity under five input conditions, with the Input and Output configurations following the tokenizer masking setup. The DriveUniWAM block uses the same input condition as the tokenizer evaluation, allowing the effect of each modality to be tracked from representation quality to downstream prediction. Because these variants mask modalities at evaluation time rather than retraining with modality dropout, we do not interpret them as a complete sensor-failure robustness study.

\begin{table*}[t]
\caption{Modality sensitivity for reconstruction and joint future generation on NAVSIM navtest. C, L, and T denote camera, LiDAR/depth, and ego-trajectory modalities. Input and reconstruction Output follow the original tokenizer masking setup; the DriveUniWAM block evaluates future generation under the same input condition. Best results are highlighted in \textbf{bold}.}
\label{tab:modality-ablation}
\centering
\scriptsize
\setlength{\tabcolsep}{3.5pt}
\renewcommand{\arraystretch}{1.18}
\resizebox{\textwidth}{!}{%
\begin{tabular}{c c c c c c c c c c}
\toprule
& & \multicolumn{4}{c}{Tokenizer Reconstruction} & \multicolumn{4}{c}{Joint Future Generation} \\
\cmidrule(lr){3-6} \cmidrule(lr){7-10}
Input & Output & LPIPS$_{\mathrm{DINO}}\downarrow$ & rFVD$_{\mathrm{VJ}}\downarrow$ & Depth AbsRel$\downarrow$ & Ego ADE$\downarrow$ & LPIPS$_{\mathrm{DINO}}\downarrow$ & gFVD$_{\mathrm{VJ}}\downarrow$ & Depth AbsRel$\downarrow$ & Traj. ADE$\downarrow$ \\
\midrule
C & C+L+T & 0.1671 & 8.8645 & 0.1274 & 1.0512 & 0.1170 & 12.1153 & 0.2185 & 3.1991 \\
L & L+T & -- & -- & 0.1202 & 0.3350 & -- & -- & 0.2934 & 4.5854 \\
T & T & -- & -- & -- & 0.2978 & -- & -- & -- & 2.3534 \\
\addlinespace[4pt]
C+L & C+L+T & 0.1070 & 3.6430 & 0.0463 & 0.0876 & 0.0985 & 8.8587 & 0.1669 & 2.6611 \\
\rowcolor{gray!10}
\textbf{C+L+T} & \textbf{C+L+T} & \textbf{0.1063} & \textbf{3.4546} & \textbf{0.0461} & \textbf{0.0728} & \textbf{0.0955} & \textbf{6.6076} & \textbf{0.1059} & \textbf{0.3816} \\
\bottomrule
\end{tabular}
}%
\end{table*}

Using all three input modalities yields the strongest overall
reconstruction and future-generation results, indicating that the unified
latent benefits from jointly integrating complementary visual, geometric,
and motion evidence.
\FloatBarrier
\section{Qualitative NAVSIM Planning Visualizations}
\label{sec:qualitative-navsim}

The qualitative comparisons cover both vision-language-action and world-action modeling paradigms. Each group contains a front-view observation followed by predictions from AutoVLA and ReCogDrive (VLA methods), PWM (a world-action baseline), DriveUniWAM (Ours), and DriveUniWAM with reinforcement learning post-training. AutoVLA produces reasonable paths in straight-road scenes with simple geometry, but can under-steer at intersections and turns, resulting in large turning radii or endpoints that deviate from the ground-truth route. ReCogDrive generally makes better use of road topology, while PWM captures the broad scene-conditioned motion trend but can retain local curvature or endpoint errors in more demanding cases. DriveUniWAM further improves trajectory shape, turning curvature, and endpoint placement. After reinforcement learning post-training, the predicted paths show greater stability in complex intersections, turns, merges, and scenes with an explicit navigation direction: the model adjusts its direction earlier and better matches the ground-truth curvature, position, and final heading. Overall, the visualizations show that DriveUniWAM produces feasible trajectories across both comparison paradigms, while reinforcement learning post-training further improves their agreement with driving feasibility.

\begin{figure*}[tb]
\centering
\includegraphics[height=0.88\textheight,keepaspectratio]{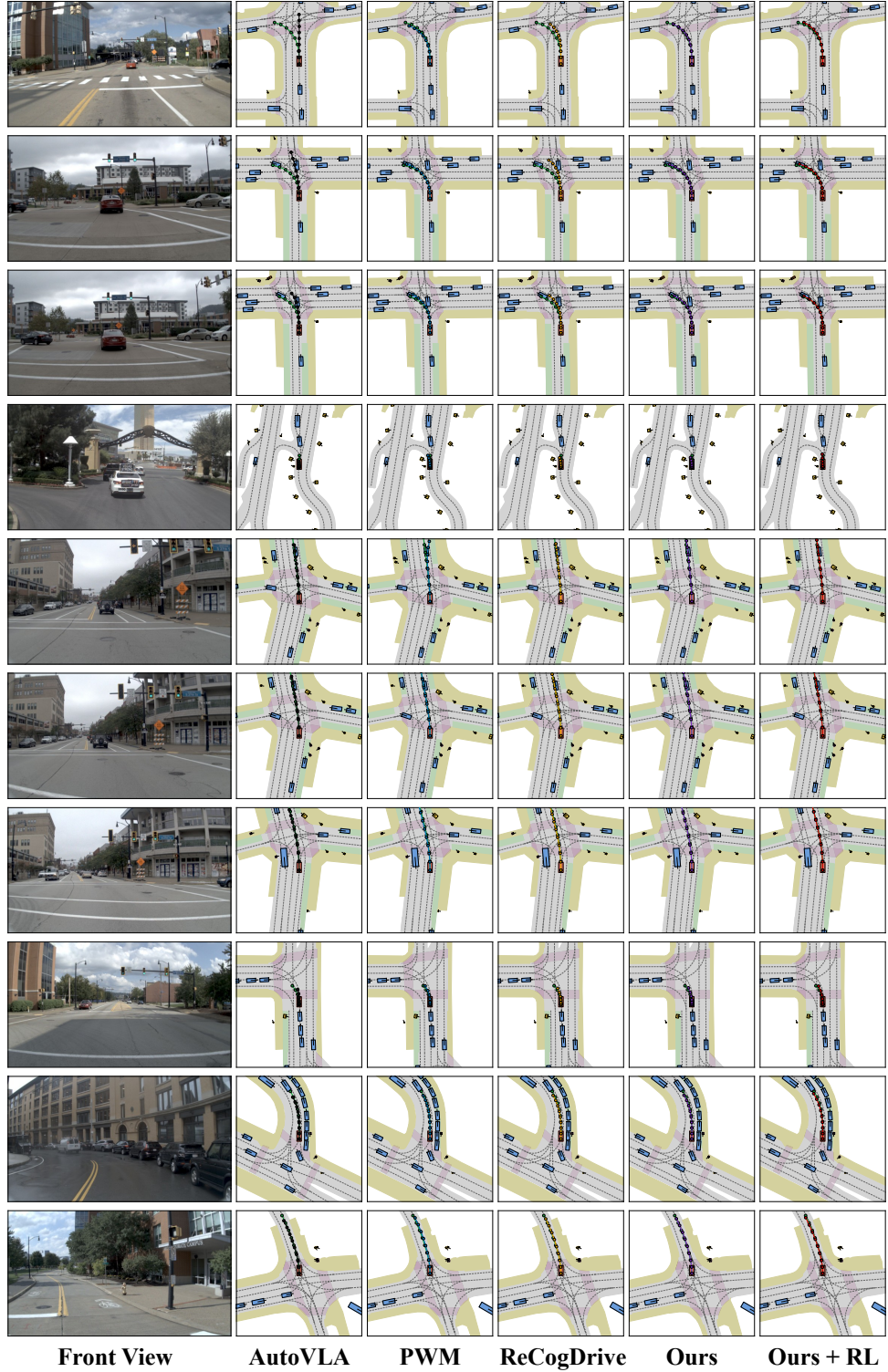}
\caption{Qualitative NAVSIM navtest visualizations for scene group 1. From left to right, the columns show the front-view observation, AutoVLA, ReCogDrive, PWM, DriveUniWAM (Ours), and DriveUniWAM with reinforcement learning post-training. AutoVLA and ReCogDrive are vision-language-action methods, whereas PWM is a world-action baseline.}
\label{fig:navtest-qualitative-1}
\end{figure*}

\begin{figure*}[tb]
\centering
\includegraphics[height=0.88\textheight,keepaspectratio]{figs/navtest_fig2.pdf}
\caption{Qualitative NAVSIM navtest visualizations for scene group 2. The columns follow the ordering defined in Fig.~\ref{fig:navtest-qualitative-1}.}
\label{fig:navtest-qualitative-2}
\end{figure*}

\begin{figure*}[tb]
\centering
\includegraphics[height=0.88\textheight,keepaspectratio]{figs/navtest_fig3.pdf}
\caption{Qualitative NAVSIM navtest visualizations for scene group 3. The columns follow the ordering defined in Fig.~\ref{fig:navtest-qualitative-1}.}
\label{fig:navtest-qualitative-3}
\end{figure*}

\begin{figure*}[tb]
\centering
\includegraphics[height=0.88\textheight,keepaspectratio]{figs/navtest_fig4.pdf}
\caption{Qualitative NAVSIM navtest visualizations for scene group 4. The columns follow the ordering defined in Fig.~\ref{fig:navtest-qualitative-1}.}
\label{fig:navtest-qualitative-4}
\end{figure*}

\begin{figure*}[tb]
\centering
\includegraphics[height=0.88\textheight,keepaspectratio]{figs/navtest_fig5.pdf}
\caption{Qualitative NAVSIM navtest visualizations for scene group 5. The columns follow the ordering defined in Fig.~\ref{fig:navtest-qualitative-1}.}
\label{fig:navtest-qualitative-5}
\end{figure*}

\begin{figure*}[tb]
\centering
\includegraphics[height=0.88\textheight,keepaspectratio]{figs/navtest_fig6.pdf}
\caption{Qualitative NAVSIM navtest visualizations for scene group 6. The columns follow the ordering defined in Fig.~\ref{fig:navtest-qualitative-1}.}
\label{fig:navtest-qualitative-6}
\end{figure*}

\begin{figure*}[tb]
\centering
\includegraphics[height=0.88\textheight,keepaspectratio]{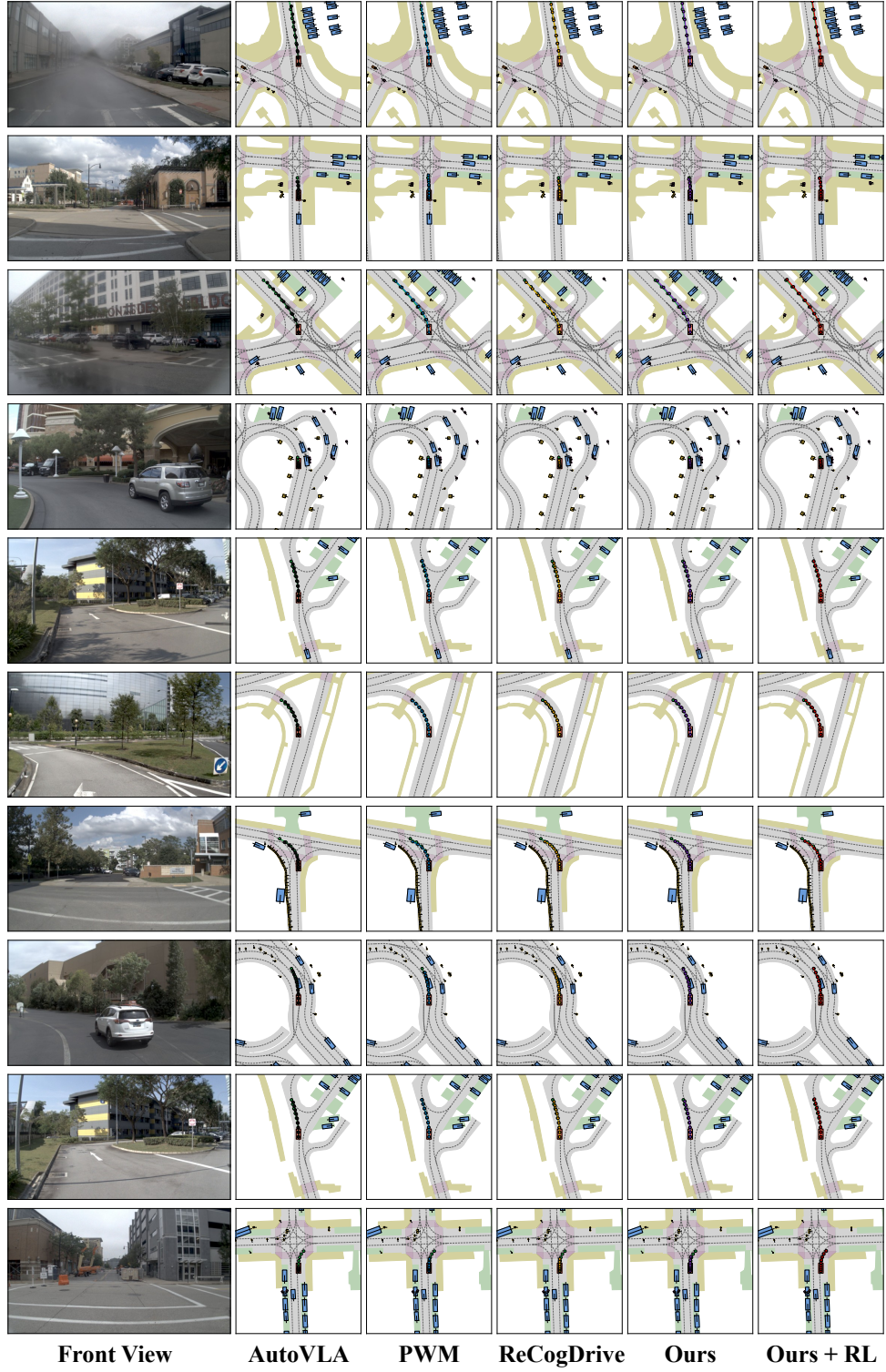}
\caption{Qualitative NAVSIM navtest visualizations for scene group 7. The columns follow the ordering defined in Fig.~\ref{fig:navtest-qualitative-1}.}
\label{fig:navtest-qualitative-7}
\end{figure*}

\begin{figure*}[tb]
\centering
\includegraphics[height=0.88\textheight,keepaspectratio]{figs/navtest_fig8.pdf}
\caption{Qualitative NAVSIM navtest visualizations for scene group 8. The columns follow the ordering defined in Fig.~\ref{fig:navtest-qualitative-1}.}
\label{fig:navtest-qualitative-8}
\end{figure*}

\begin{figure*}[tb]
\centering
\includegraphics[height=0.88\textheight,keepaspectratio]{figs/navtest_fig9.pdf}
\caption{Qualitative NAVSIM navtest visualizations for scene group 9. The columns follow the ordering defined in Fig.~\ref{fig:navtest-qualitative-1}.}
\label{fig:navtest-qualitative-9}
\end{figure*}

\begin{figure*}[tb]
\centering
\includegraphics[height=0.88\textheight,keepaspectratio]{figs/navtest_fig10.pdf}
\caption{Qualitative NAVSIM navtest visualizations for scene group 10. The columns follow the ordering defined in Fig.~\ref{fig:navtest-qualitative-1}.}
\label{fig:navtest-qualitative-10}
\end{figure*}